\documentclass[10pt,twocolumn,letterpaper]{article}

\usepackage{iccv}
\usepackage{times}
\usepackage{epsfig}
\usepackage{graphicx}
\usepackage{amsmath}
\usepackage{amssymb}

\usepackage{authblk}

\usepackage{color}
\usepackage{algorithm}
\usepackage{algpseudocode}
\usepackage{dsfont}
\usepackage{subcaption}
\usepackage{cite}
\usepackage{booktabs}
\usepackage{tabularx}
\usepackage{url}
\usepackage{soul}
\usepackage{balance}
\usepackage{paralist} 
\usepackage{lscape}
\usepackage{array,booktabs}
\usepackage{subcaption}
\usepackage{rotating}

\usepackage{caption}
\usepackage{subcaption}
\usepackage[accsupp]{axessibility}

\newcommand{\rCircNum}{\textcolor{red}{{\normalsize \textcircled{\footnotesize{1}}}}}
\newcommand{\gCircNum}{\textcolor{green}{{\normalsize \textcircled{\footnotesize{2}}}}}
\newcommand{\bCircNum}{\textcolor{blue}{{\normalsize \textcircled{\footnotesize{3}}}}}

\newcommand{\yan}[1]{\textcolor{black}{#1}}
\newcommand{\song}[1]{\textcolor{black}{#1}}

\usepackage[breaklinks=true,bookmarks=false]{hyperref}

\iccvfinalcopy 

\def\iccvPaperID{6872} 

\ificcvfinal\fi

\begin{document}

\title{DepthTrack: Unveiling the Power of RGBD Tracking}


\author{Song Yan$^{1, \dagger}$, 
Jinyu Yang$^{2, 3, \dagger}$, 
Jani K\"apyl\"a$^{1, \dagger}$,
Feng Zheng$^2$,
Ale\v{s} Leonardis$^3$, 
Joni-Kristian K\"am\"ar\"ainen$^1$\\
$^1$Tampere University ~~~
$^2$Southern University of Science and Technology ~~~
$^3$University of Birmingham \\
{\tt\small \{song.yan,jani.kapyla,joni.kamarainen\}@tuni.fi, 
jinyu.yang96@outlook.com,
zhengf@sustech.edu.cn,
a.leonardis@cs.bham.ac.uk}
}

\maketitle
\ificcvfinal\thispagestyle{empty}\fi

\let\thefootnote\relax\footnotetext{\textsuperscript{$\dagger$}Equal contribution.}

\begin{abstract}
RGBD (RGB plus depth) object tracking is gaining momentum as RGBD sensors have become popular in many application fields such as robotics. 
However, the best RGBD trackers are extensions of the state-of-the-art deep RGB trackers. 
They are trained with RGB data and the depth channel is used as a sidekick for subtleties such as occlusion detection.
This can be explained by the fact that there are no sufficiently large RGBD datasets to 1) train ``deep depth trackers'' and to 2) challenge RGB trackers with sequences for which the depth cue is essential.
This work introduces a new RGBD tracking dataset - DepthTrack - that has twice as many sequences (200) and scene types (40) than in the largest existing dataset, and three times more objects (90). 
In addition, the average length of the sequences (1473), the number of deformable objects (16) and the number of annotated tracking attributes (15) have been increased.
Furthermore, by running the SotA RGB and RGBD trackers on DepthTrack, we propose a new RGBD tracking baseline, namely DeT, which reveals that deep RGBD tracking indeed benefits from genuine training data.
The code and dataset is available at \url{https://github.com/xiaozai/DeT}.
\end{abstract}

\section{Introduction}
%
The goal of generic object tracking is to localize an unknown object in a video sequence given
its position in the first frame. The most popular tracking modality is color vision in the terms of RGB image frames as the input. Furthermore, the problem of
visual object tracking (VOT) can be divided into short-term and long-term tracking  which are evaluated
differently~\cite{kristan2019seventh,Kristan2020a}. The short-term evaluation protocol focuses on
the tracker itself by measuring its accuracy and robustness. If a tracker loses the target, it affects to the robustness metric, but the tracker is re-initialized and then evaluation continues. In the long-term (LT) protocol the tracker is not re-initialized and thus the LT trackers need special procedures for detecting whether the target is present or not and re-detection in the target-not-present operation mode.
The two latest VOT challenges~\cite{kristan2019seventh,Kristan2020a} include additional tracks for RGBT (RGB plus Thermal infrared) and RGBD (RGB plus Depth) object tracking. Interestingly, there are no trackers specialized on
thermal features or depth features, but the top performing RGBT and RGBD trackers all use RGB features
learned by the leading deep tracker architectures, MDNet~\cite{MDNet}, ATOM~\cite{2019ATOM} and DiMP~\cite{bhat2019learning}.
The additional modality T or D is used only as a ``sidekick'' to help in various special cases
such as occlusion detection. Therefore it remains unclear what are the potential applications of
RGBD and RGBT tracking and whether T and D channels have their own powerful features.

\begin{figure}
\centering
\includegraphics[width=1.05\linewidth]{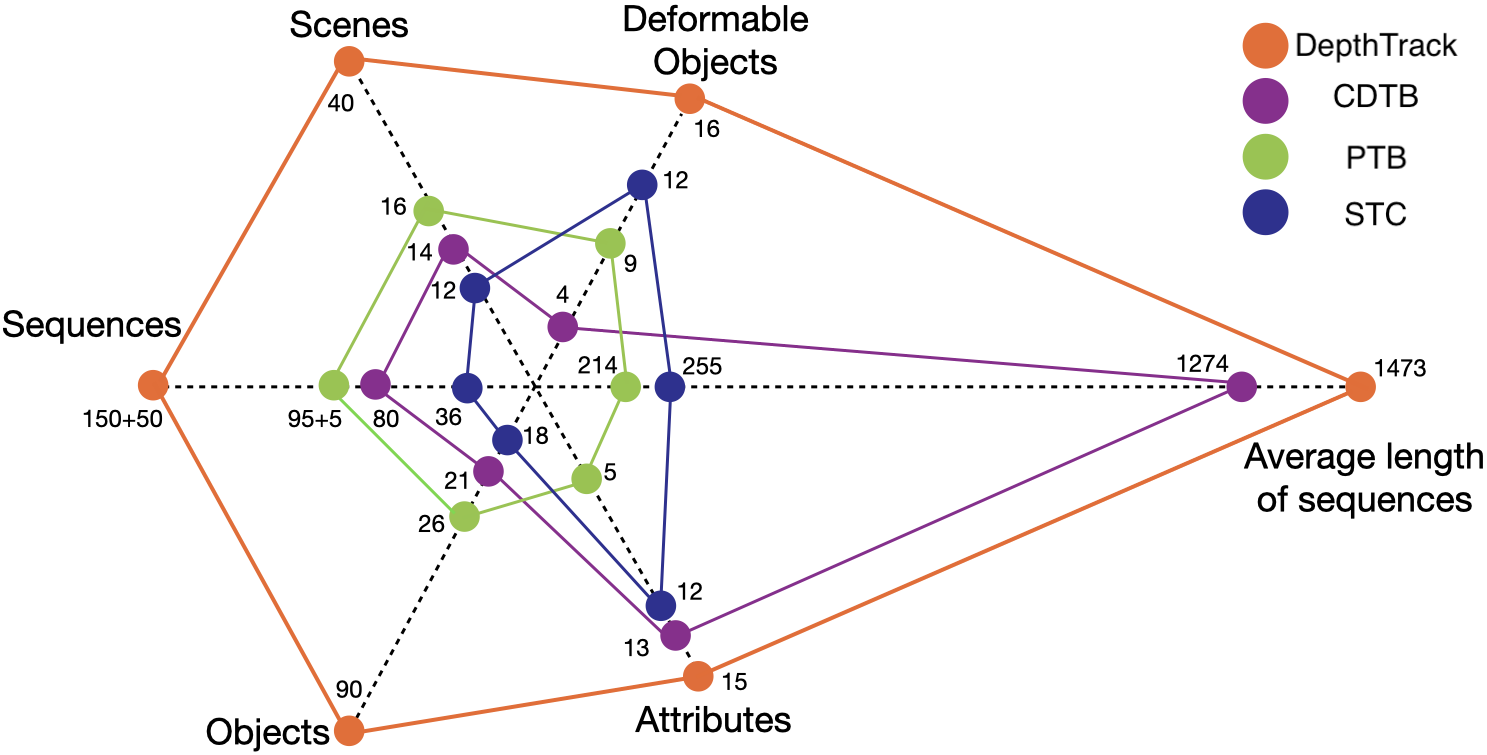}
\caption{Comparison of the properties of the proposed DepthTrack and the previous RGBD tracking datasets.}
\label{fig:fig_dataset_statistic}
\end{figure}

In this work we focus on RGBD tracking. 
RGBD tracking has been investigated in a number of recent works
(\textit{e.g.},~\cite{liu2018context,Kart_ECCVW_2018,Kart_CVPR_2019}), but these use
conventional RGB and depth features and are inferior to the deep RGB tracker based methods in the VOT 2019 and 2020 RGBD
challenges~\cite{kristan2019seventh,Kristan2020a}. The top performing
trackers in VOT 2020 RGBD challenge, ATCAIS, DDiMP and CLGS\_D, are based on the
recent deep RGB trackers, ATOM~\cite{2019ATOM}, DiMP~\cite{bhat2019learning} and MDNet~\cite{MDNet},
and use depth only to detect occlusion, target disappearance and target re-detection. There are no ``depth trackers'' that are trained with depth sequences. The main reason is the lack of suitable training data. For example, the three existing RGBD datasets, PTB~\cite{song2013tracking}, STC~\cite{xiao2017robust} and CDTB~\cite{Lukezic_2019_ICCV}, contain only 100+36+80 sequences. The target and attribute annotations are available only for the STC and CDTB datasets leaving only 116 RGBD tracking sequences for research purposes. At the same time, the existing RGB tracking datasets, LaSOT\cite{fan2019lasot},
TrackingNet~\cite{muller2018trackingnet} and GOT-10k~\cite{Huang_2019}, contain 1.55K+30K+10K sequences, \textit{i.e.} there is more than two orders of magnitude difference in the amount of data. To unveil the power of depth in RGBD tracking, we need larger and more diverse RGBD or depth-only tracking datasets.

\begin{figure*}[h]
\centering
\includegraphics[width=0.99\linewidth]{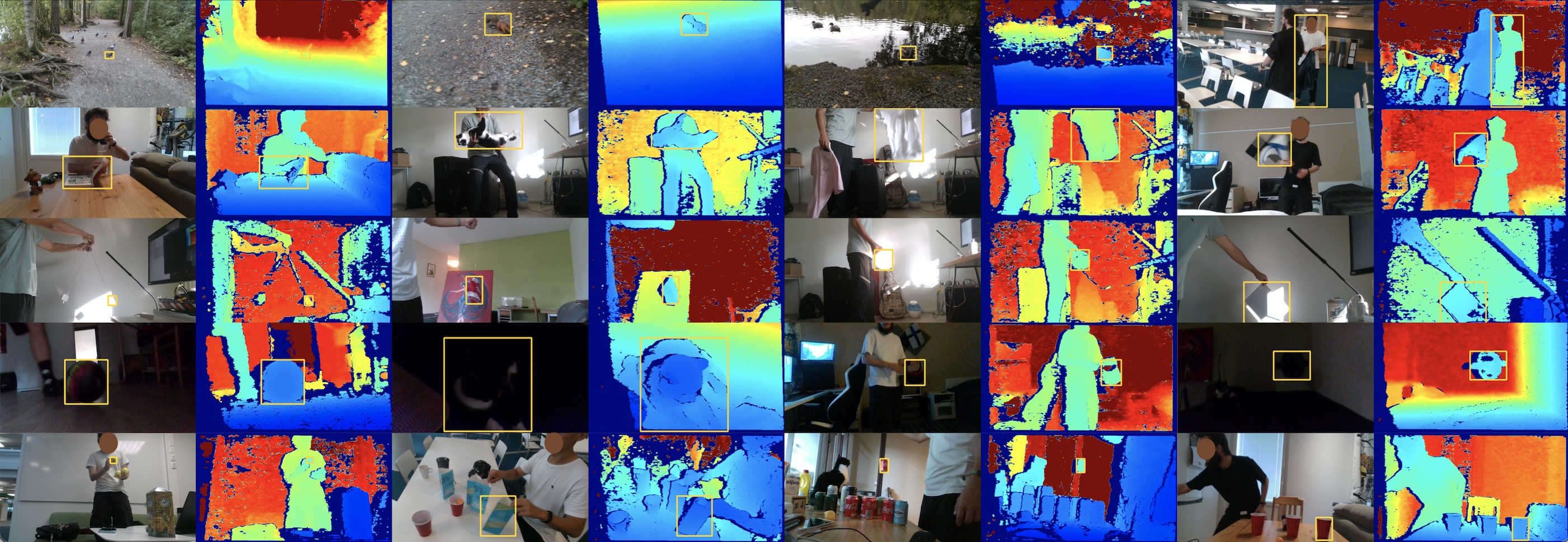}
\caption{Sample frames from DepthTrack (targets marked with yellow boxes). Some of these sequences are particularly difficult for RGB only: animals and humans that camouflage to the background (1st row),
deformable objects (2n row);
similar target and background color (3rd row),
dark scenes (4th row) and multiple similar objects (5th row).}
\label{fig:fig_dataset_intro}
\end{figure*}

In this work, we report a new fully annotated RGBD dataset which is the largest and most diverse of the available datasets in this area. 
The main contributions are:
\begin{compactitem}
    \item A new annotated dataset - DepthTrack - for RGBD tracking. The dataset is substantially larger than its predecessors and is divided to training and test sets (Fig.~\ref{fig:fig_dataset_statistic} and Fig.~\ref{fig:fig_dataset_intro}).
    \item Results and findings from extensive experiments with the SotA RGB and RGBD trackers on DepthTrack. 
    These findings, including the fusion of RGB and depth features, will facilitate future works on collecting better RGBD tracking datasets and developing better RGBD trackers.
    \item A new RGBD baseline, DeT, that is trained with depth tracking data and obtains better RGBD tracking performance than the existing SotA trackers.
\end{compactitem}
DepthTrack RGBD sequences, annotation meta data and evaluation code are made compatible with the VOT 2020 Python Toolkit to make it easy to evaluate existing and new trackers with DepthTrack.

\section{Related Work}
\paragraph{RGBD tracking datasets.}
There are only three publicly available datasets for
RGBD object tracking:
{\bf 1)} \textit{Princeton Tracking Benchmark} (PTB)~\cite{song2013tracking},
{\bf 2)} \textit{Spatio-Temporal Consistency} dataset (STC)~\cite{xiao2017robust}
and
{\bf 3)} \textit{Color and Depth Tracking Benchmark} (CDTB)~\cite{Lukezic_2019_ICCV}.
The statistics of these datasets are compared to the proposed DepthTrack in
Fig.~\ref{fig:fig_dataset_statistic}.
PTB contains 100 RGBD video sequences of rigid and nonrigid objects recorded with Kinect in indoors. The dataset diversity is rather limited in the number of scenes and attributes and approximately 14\% of sequences have RGB and D syncing errors and 8\% are miss-aligned.
STC addresses the drawbacks of PTB. The dataset is recorded by Asus Xtion RGBD sensor and contains mainly indoor sequences and a small number of
low-light outside sequences. The dataset is smaller than PTB, containing only 36 sequences, but contains annotations of thirteen attributes. STC addresses short-term tracking.
CDTB is the most recent dataset and is utilized in the VOT-RGBD 2019 and
2020 challenges \cite{kristan2019seventh, Kristan2020a}.
It contains 80 sequences and 101,956 frames in total. The sequences are
recorded both indoors and outdoors and the authors use evaluation
protocols from the VOT challenge.

\paragraph{RGBD tracking algorithms.}
Until very recently the RGBD trackers were based on
engineered features and used various ad hoc methods to
combine RGB and D~\cite{liu2018context,Kart_ECCVW_2018,Kart_CVPR_2019},
for example, DAL\cite{2019DAL} embedded the depth information into RGB deep features through the reformulation of deep discriminative correlation filter.
Furthermore, the best performing trackers in VOT-RGBD2019 and VOT-RGBD2020
challenges are extensions of well-known deep RGB trackers~\cite{kristan2019seventh,Kristan2020a}. For example,
the three winning RGBD trackers in VOT-RGBD2020 are ATCAIS, DDiMP and CLGS\_D, which are extensions of the deep RGB trackers
ATOM~\cite{2019ATOM}, DiMP~\cite{bhat2019learning} and MDNet~\cite{MDNet},
respectively. The main tracking cue for these trackers is RGB and
depth is used only for long-term tracking tasks. In this work,
we report results for the first RGBD tracker whose depth branch
is trained with genuine depth data and that provides superior results.

\section{The DepthTrack Dataset}
%

\begin{figure*}[h]
\centering
\includegraphics[width=0.99\linewidth]{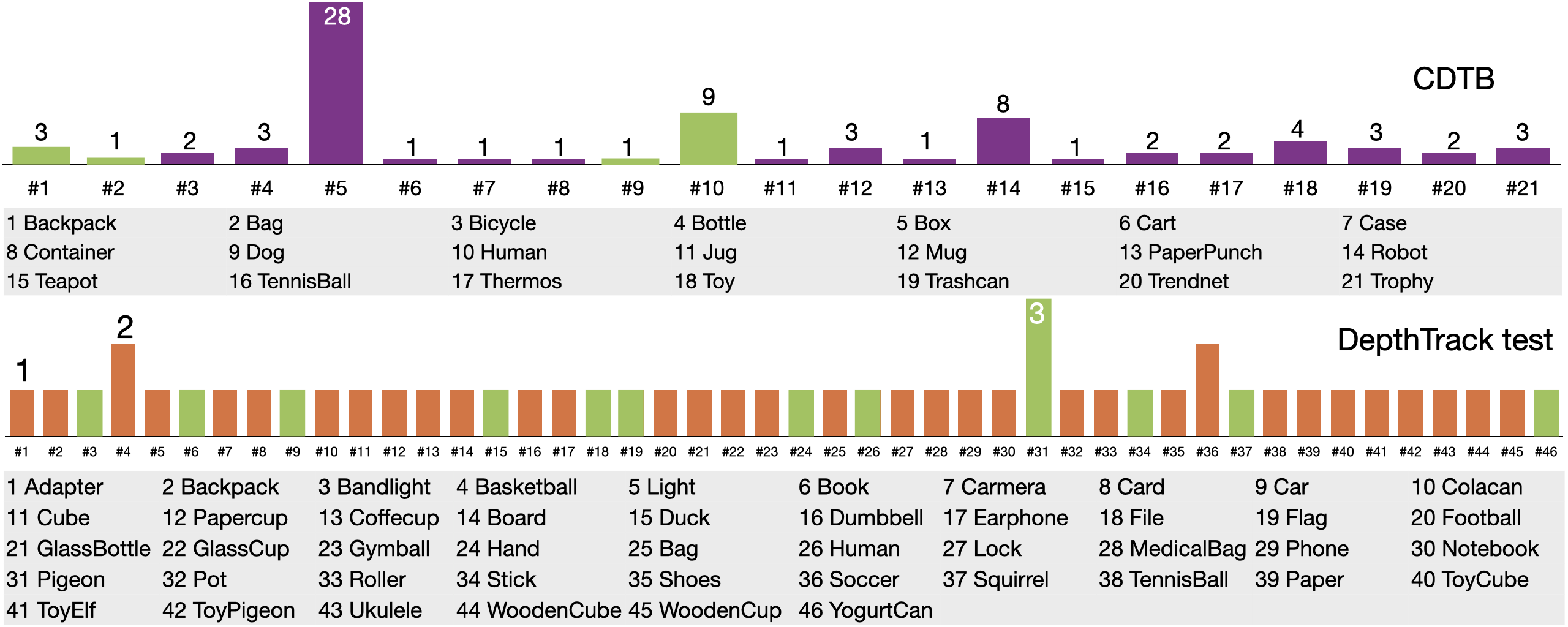}
\caption{Distributions of different target types in CDTB~\cite{Lukezic_2019_ICCV} (top) and the DepthTrack test set of 50 sequences (bottom). Green color denotes non-rigid objects. 
Purple and orange denote the rigid objects. 
}
\label{fig:fig_dataset_object_types_distribution}
\end{figure*}                                 


\paragraph{Tracking sequences.}
CDTB dataset~\cite{Lukezic_2019_ICCV} was captured with multiple
active (Kinect-type) and passive (stereo) depth sensors and RGB
cameras, but in the VOT evaluations the acquisition devices had only marginal
effect to the results while the sequence content was the main factor. Therefore
DepthTrack was captured with a single mid-price but high quality RGBD sensor:
\textit{Intel Realsense 415}.
The RGB and depth frames were stored using the same
$640 \times 360$ resolution and the frame rate of 30~fps.
RGB images were stored as 24-bit JPEG with low compression rate
and the depth frames as 16-bit PNG.
The Intel sensor provides the RGB and depth frames temporally synchronized.

In our data collection we particularly focused on content
diversity.
The overall properties of DepthTrack are compared to three available datasets in Fig.~\ref{fig:fig_dataset_statistic} and a more detailed comparison on different target types between DepthTrack
and CDTB is shown in 
Fig.~\ref{fig:fig_dataset_object_types_distribution}.
It is notable that there are three object types that dominate the CDTB sequences: ``box'', ``human'' and ``robot''. At the same time,
the high diversity of DepthTrack provides unique targets in almost every test sequence.
Another important factor is presence of humans in
CDTB sequences and them moving simple rigid targets objects. To reduce
the ``human operator bias'' in DepthTrack we included a large number of
\yan{objects that are only indirectly manipulated by hand, \eg by a barely visible rope.}

DepthTrack is split to 150 training sequences and
50 test sequences so that they do not contain common scenes
and objects. The training set contains 218,201 video frames
and the test set 76,390. As shown in
Fig.~\ref{fig:fig_dataset_object_types_distribution} almost
every test sequence has its own unique target type. Only ``basketball''
(2), ``soccer'' (2) and ``pigeon'' (3) appear multiple times in the test set.

\paragraph{Data annotation.}
Each DepthTrack frame is annotated with the target bounding box location and multiple scene attributes that help to analyze the results.
The axis-aligned bounding box annotation is adopted from the VOT-RGBD challenges
and we followed the VOT annotation protocol and employed the Aibu\textsuperscript{1} annotation tool.
\footnotetext{\textsuperscript{1}https://github.com/votchallenge/aibu}
\yan{In the protocol the axis-aligned bounding box should tightly fit the target object in each frame to avoid the background pixels.}

To allow detailed scene-type level analysis of tracking results we annotated 15 per-frame attributes.
Besides the 13 attributes used in CDTB (VOT-RGBD 2019 and 2020), we introduced two new attributes: {\em background clutter}~(BC) and {\em camera motion}~(CM). BC denotes scenes where the target and background share the same color or texture and CM denotes cases where camera movement 
\yan{leads to} substantial target distance (depth) change. BC frames are expected to be difficult for RGB-only tracking and CM frames for D-only tracking.
For the detailed description of each attribute, please refer to our supplementary material.
The total time to annotate the DepthTrack data was more than 150 hours.

\paragraph{Performance metrics.}
DepthTrack is a long-term RGBD tracking dataset where the trackers must detect
when targets are not visible and again re-detect them when they become visible
again. Our evaluation protocol is adopted from~\cite{Lukezic_2019_ICCV}.

A single pass of each sequence is made and for each frame of the sequence 
trackers must provide a target visibility {\em confidence score} and
{\em bounding box coordinates}. The bounding box coordinates are
used to evaluate their precision with respect to the ground truth bounding
boxes. Precision is measured by the bounding box overlap ratio. Confidence
scores are used to evaluate whether the tracker is able to recall the frames
where the targets are annotated visible. Ideally confidence is $0$ for the
frames without target and $1$ for the frames where the target or parts of it
are visible.

The overall evaluation is based on the tracking precision ($Pr$) and recall ($Re$) metrics~\cite{kristan2019seventh}. 
Tracking precision is the measure of target localization accuracy when the target
is labelled visible. 
\yan{Tracking recall measures the accuracy of classifying the labelled visible target.} 
As a single measure, the \textit{F-score} is used as the harmonic mean of precision and recall to rank
the trackers. The precision and recall are computed for each frame and then
averaged over all $j=1,\ldots,N_i$ frames of the $i$-th sequence. That provides per
sequence metrics which are then averaged over all $i=1,\ldots,N$ sequences
to compute dataset specific metrics. It should be noted, that the tracker
confidence affects to \textit{Pr} and \textit{Re} and therefore the {\em precision-recall graphs}
are computed by varying the confidence threshold $\tau$.

One weakness of the above averaging protocol is that if the video lengths vary substantially,
then the short videos may get unreasonably large weight in the performance metrics. For example, the sequence length  varies between 143 to 3816 frames in DepthTrack. 
To alleviate this problem, we do averaging over all frames of all sequences (refer as \textbf{frame-based} evaluation), as well as the sequence specific averaging (refer as \textbf{sequence-based} evaluation).
As the final performance metric, we store the highest F-score over all confidence thresholds $\tau$ and store also the corresponding precision and recall values. For the details of our evaluation metrics, please refer to the supplementary material.

\section{Baseline RGBD Trackers}
In this section we introduce the baseline RGBD trackers used
in our experiments. The existing baseline trackers are selected
among the best performing RGB and RGBD trackers in the
recent VOT-RGBD evaluations~\cite{kristan2019seventh,Kristan2020a}.
In the spirit of our work, we introduce a new depth data trained baseline in
Section~\ref{sec:our_tracker}.

\subsection{\song{Existing Baselines}}
\label{sec:existing_baselines}
To establish a strong set of baseline trackers we selected the following 23 trackers:
\begin{compactitem}
\item 7 winning submissions to the VOT-RGBD 2019 and 2020 challenges~\cite{kristan2019seventh,Kristan2020a}:
ATCAIS, DDiMP, CLGS\_D, SiamDW\_D, LTDSEd, Siam\_LTD and SiamM\_Ds;
\item 3 RGBD baselines using hand-crafted RGB and depth features: 
DS-KCF~\cite{camplani2015real}, DS-KCF-Shape~\cite{2016DS} and CA3dMS~\cite{liu2018context};
\item A recent RGBD tracker that embeds D cue into the RGB channels: DAL~\cite{2019DAL};
\item 3 winning submissions to the RGB VOT-LT (long-term) 2020 challenge: 
RLT\_DiMP~\cite{choi2020robust}, LTMU\_B, Megatrack;
\item 3 winning submissions to the VOT-ST (short-term) 2020 challenge: 
RPT~\cite{ma2020rpt}, OceanPlus and AlphaRef~\cite{yan2020alpha};
\item 6 deep RGB trackers used by the best RGBD trackers: 
SiamFC~\cite{bertinetto2016fully}, SiamRPN~\cite{2018High}, ATOM~\cite{2019ATOM}, DiMP50~\cite{bhat2019learning}, D3S~\cite{lukezic2020d3s} and PrDiMP50~\cite{2020Know}.
\end{compactitem}

\noindent
All trackers use the code of their original authors and
their default parameter settings. Note that the original authors have
tuned the parameters for the three existing datasets
STC, PTB and CDTB. In particular, the RGBD trackers submitted to
VOT are optimized for CDTB.

\begin{figure}[t]
\centering
\includegraphics[width=0.95\linewidth]{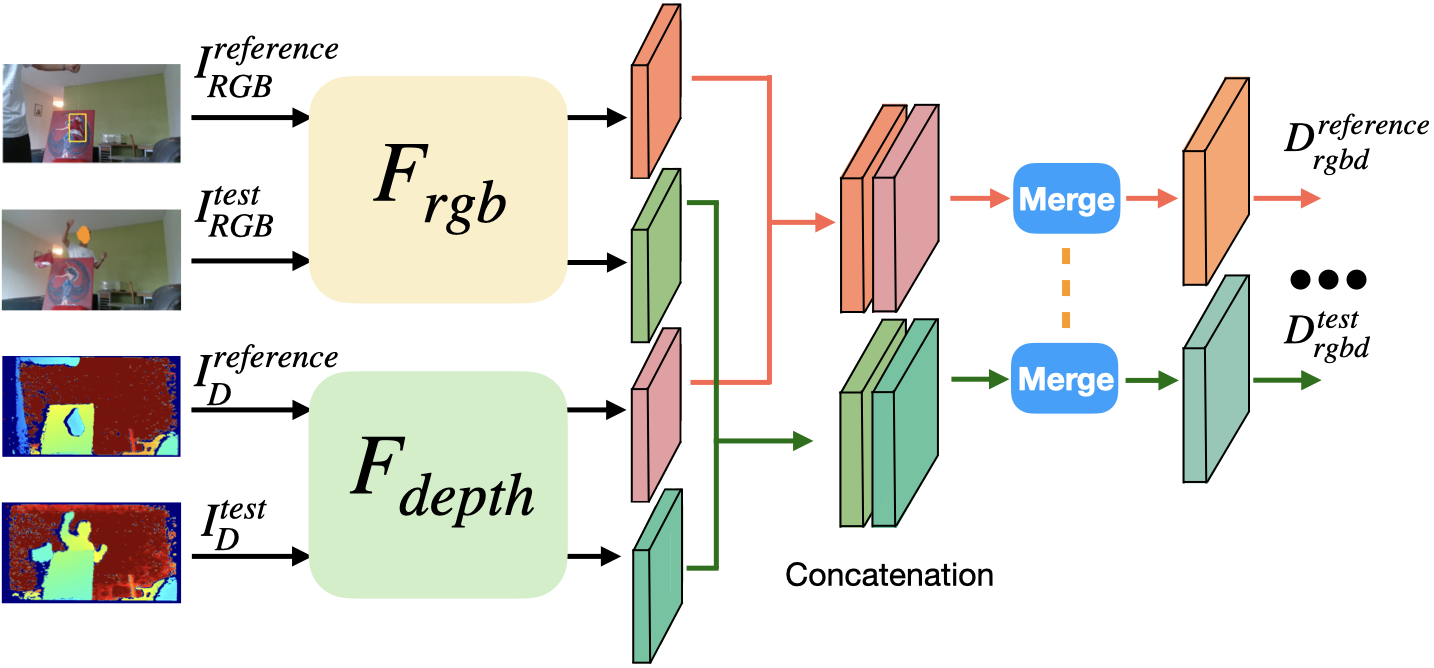}
\caption{The feature extraction and merging part of the proposed DeT RGBD tracker. 
The tracking head depends on the selected architecture (e.g. ATOM or DiMP) and can be changed.
$F$-s are the backbone networks (e.g. ResNet), 
but in our case the D backbone branch is trained from scratch with depth tracking data 
and thus unveils the power of depth for tracking.
}
\label{fig:trainable_rgbd_tracker}
\end{figure} 
%
\subsection{A New Baseline - DeT}
\label{sec:our_tracker}
We propose a new RGBD baseline for the DepthTrack
dataset - {\em DepthTrack Tracker (DeT)} - that obtains the best results thanks to deep depth features learned from depth data.
The tracker architecture (Fig.~\ref{fig:trainable_rgbd_tracker}) is inspired by the online structures of the recent SotA
trackers ATOM and DiMP. 
\yan{Therefore, our baseline is actually the depth feature extractor and a feature pooling layer and can be combined with either ATOM or DiMP tracker head that performs the actual tracking.}
\yan{The main difference of the DeT and the original ATOM or DiMP}
is that DeT extracts more powerful depth features learned from depth tracking data.

\paragraph{Generating depth tracking data.}
The main problem in deep RGBD tracking is the lack
of depth data as the existing datasets provide only
116 sequences with RGB and D images (note that the PTB
annotations are not public). Therefore, we developed
a simple procedure to generate depth data for the
existing large RGB tracking datasets.
We employed the monocular depth estimation algorithm DenseDepth~\cite{Alhashim2018} on the existing RGB benchmarks LaSOT~\cite{fan2019lasot} and COCO~\cite{lin2014microsoft}, and manually
selected the best 646 sequences from LaSOT
and 82K images from COCO.
These estimated depth images are used to pre-train DeT
from scratch.

\paragraph{RGBD features for tracking.}
The generated depth tracking sequences
are used to train the DeT tracker from scratch
and using similar offline training as used for
ATOM and DiMP. Similar to RGB data in the original works, the training takes 50 epochs after which the training error does not improve anymore. After that the RGB trained color features and D(epth) trained depth features
are extracted from the two feature paths as
$\left\{D_{RGB},D_{D}\right\} = \left\{F_{rgb}(I_{RGB}),F_{depth}(I_{D})\right\}$
and are computed separately for the reference and test branches
of the master tracker: $D^{ref}$ and $D^{test}$.
Since the master trackers, ATOM and DiMP, require a single feature
images the RGB and D channels need to be merged to
$D^{ref}_{RGBD}$ and $D^{test}_{RGBD}$.

We want to learn the RGB and D merge from training data and
thus adopt a pooling layer as the standard solution.
In our preliminary experiments,
we compared a number of the typical pooling operations and found the element-wise maximum operation to perform the best.
After the pooling operation, 
\yan{the two outputs of the DeT feature extraction part in}
Fig.~\ref{fig:trainable_rgbd_tracker} represent the
{\em reference} (the previous frame with an estimated bounding box)
and {\em test} (the current frame) branches in the DiMP and ATOM terminology 
\yan{and the tracking head}
depends on whether ATOM,
DiMP or another similar online tracking part is used.

\section{Experiments}
\label{sec:experiments}
All experiments were run on the same computer with an Intel i7- CPU @ 3.40GHZ, 16 GB RAM, and one NVIDIA GeForce RTX 3090 GPU.
In the experiments, we adopt the DiMP50 as the master tracker and employ the element-wise maximum operation as the feature pooling layer, and we refer it as DeT for short.
\yan{DeT variants adopt the same backbones as the master trackers, \eg ResNet50 for DiMP50 or ResNet18 for ATOM.}

\newcolumntype{C}{>{\centering\arraybackslash}X}
\begin{table*}[t]
\begin{center}
\caption{
Tracking results for the DepthTrack test set. Performance
metrics are reported for the both sequence and frame based
evaluation protocols. DeT is the proposed baseline (Sec.~\ref{sec:our_tracker})
and the input modalities (RGB/D) and tracker type (Short-Term/Long-Term)
are listed. The best three methods for each metric are marked with their rank.
}
\label{tbl:experiments_on_DeTrack}
\small
\setlength\tabcolsep{7.5pt} 
\resizebox{0.85\linewidth}{!}{%
\begin{tabularx}{\linewidth}{l | r r r | r r r | r | c c}
\toprule
~        & \multicolumn{3}{c|}{Sequence-based} & \multicolumn{3}{c|}{Frame-based}  & Speed & \multicolumn{2}{c}{Category}\\
Tracker  & \multicolumn{1}{c}{Pr} &\multicolumn{1}{c}{Re} & \multicolumn{1}{c|}{F-score} & \multicolumn{1}{c}{Pr} &\multicolumn{1}{c}{Re} & \multicolumn{1}{c|}{F-score} & \multicolumn{1}{c|}{fps} &  \multicolumn{1}{c}{LT/ST} &  \multicolumn{1}{c}{RGB/~D}\\
\midrule
DDiMP \cite{Kristan2020a}           & 0.503 & \gCircNum~0.469 & \gCircNum~0.485 & 0.490 & 0.458 & \bCircNum~0.474 & 4.77 & ST & RGBD\\
ATCAIS \cite{Kristan2020a}          & 0.500 & \bCircNum~0.455 & \bCircNum~0.476 & 0.496 & \gCircNum~0.470 & \gCircNum~0.483 & 1.32 & LT & RGBD\\
CLGS\_D \cite{Kristan2020a}         & \rCircNum~0.584 & 0.369 & 0.453 & \gCircNum~0.534 & 0.376 & 0.441 & 7.27 & LT & RGBD\\
SiamDW\_D \cite{kristan2019seventh} & 0.429 & 0.436 & 0.432 & 0.420 & 0.429 & 0.424 & 3.77 & LT & RGBD\\
LTDSEd \cite{kristan2019seventh}    & 0.430 & 0.382 & 0.405 & 0.425 & 0.383 & 0.403 & 5.65 & LT & RGBD\\
Siam\_LTD \cite{Kristan2020a}       & 0.418 & 0.342 & 0.376 & 0.410 & 0.306 & 0.350 & 13.00 & LT & RGBD\\
SiamM\_Ds \cite{kristan2019seventh} & 0.463 & 0.264 & 0.336 & 0.415 & 0.245 & 0.308 & 19.35 & LT & RGBD\\
DAL \cite{2019DAL}                  & 0.512 & 0.369 & 0.429 & 0.496 & 0.385 & 0.433 & 25.98 & LT & RGBD\\
CA3DMS \cite{liu2018context}        & 0.218 & 0.228 & 0.223 & 0.211 & 0.221 & 0.216 & 47.70 & LT & RGBD\\
DS-KCF  \cite{camplani2015real}     & 0.075 & 0.077 & 0.076 & 0.074 & 0.076 & 0.075 & 4.16  & ST & RGBD\\
DS-KCF-Shape \cite{2016DS}          & 0.070 & 0.071 & 0.071 & 0.070 & 0.072 & 0.071 & 9.47 & ST & RGBD\\
LTMU\_B \cite{Kristan2020a}        & 0.512 & 0.417 & 0.460 & 0.516 & 0.429 & 0.469  & 4.16 & LT & RGB\\
RLT\_DiMP \cite{Kristan2020a}      & 0.471 & 0.448 & 0.459 & 0.463 & 0.453 & 0.458 & 10.21 & LT & RGB\\
Megtrack \cite{Kristan2020a}       & \gCircNum~0.583 & 0.322 & 0.415 & \rCircNum~0.562 & 0.327 & 0.413 & 3.51 & LT & RGB\\
PrDiMP50 \cite{2020Know}           & 0.397 & 0.414 & 0.405 & 0.405 & 0.422 & 0.414 & 26.49 & ST & RGB\\
DiMP50 \cite{bhat2019learning}     & 0.396 & 0.415 & 0.405 & 0.387 & 0.403 & 0.395 & 42.11 & ST & RGB\\
ATOM \cite{2019ATOM}               & 0.354 & 0.371 & 0.363 & 0.329 & 0.343 & 0.336 & 43.21 & ST & RGB\\
SiamRPN \cite{2018High}            & 0.396 & 0.293 & 0.337 & 0.307 & 0.320 & 0.313 & 13.80 & ST & RGB\\
D3S \cite{lukezic2020d3s}          & 0.265 & 0.276 & 0.270 & 0.246 & 0.257 & 0.251 & 28.60 & ST & RGB\\
SiamFC \cite{bertinetto2016fully}  & 0.187 & 0.193 & 0.190 & 0.173 & 0.180 & 0.177 & 114.06 & ST & RGB\\
AlphaRef \cite{Kristan2020a}       & 0.426 & 0.443 & 0.435 & 0.448 & \bCircNum~0.467 & 0.457 & 11.01 & ST & RGB\\
RPT \cite{Kristan2020a}            & 0.436 & 0.381 & 0.406 & 0.398 & 0.369 & 0.383 & 5.75 & ST & RGB\\
OceanPlus \cite{Kristan2020a}      & 0.410 & 0.335 & 0.368 & 0.378 & 0.338 & 0.357 & 19.38 & ST & RGB\\
\midrule
\textbf{DeT}         & \bCircNum~0.560 & \rCircNum~0.506 & \rCircNum~0.532 & \bCircNum~0.523 & \rCircNum~0.479 & \rCircNum~0.500 & 36.88 & ST & RGBD \\
\bottomrule
\end{tabularx}%
}
\end{center}
\end{table*}

\subsection{Performance Evaluation}
\begin{figure}[h]
    \centering
    \includegraphics[width=\linewidth]{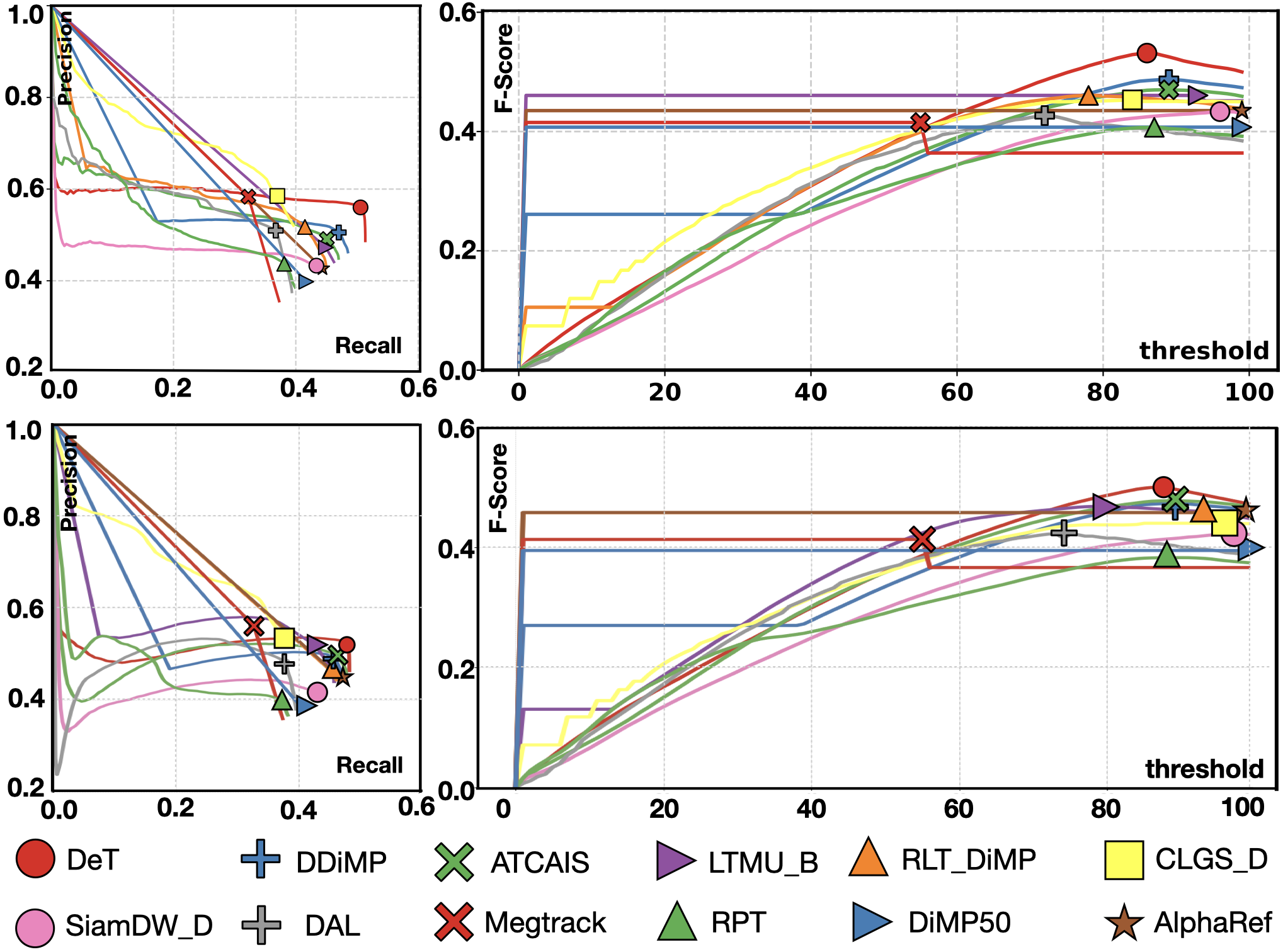}
    \caption{
    \yan{The Precision-Recall and F-score curves of the best 10 evaluated trackers ranked by their F-scores and the proposed DeT tracker and its master tracker DiMP50, 
    the best F-score point marked in each graph.
    Top: sequence-based evaluation; Bottom: frame-based evaluation.}
    }
    \label{fig:fig_trackers_pr_re_fscore_final}
\end{figure}

\paragraph{Quantitative results.}
The performance metrics and method rankings are shown in Table~\ref{tbl:experiments_on_DeTrack}. 
According to the VOT protocol the best tracker is selected based on its F-score, but precision and recall provide more details about its
performance.
\yan{Fig.~\ref{fig:fig_trackers_pr_re_fscore_final} shows the Precision-Recall and F-score plots for the proposed depth-trained DeT tracker introduced in Section~\ref{sec:our_tracker} and the best 10 SotA and baseline trackers in Section~\ref{sec:existing_baselines} ranked by their F-scores
and the RGB master tracker DiMP50~\cite{bhat2019learning}. 
Plots for all test trackers are in the supplementary material.}

The results provide the following important findings:
{\bf 1)} the VOT-RGBD2020 (using CDTB) winners DDiMP, ATCAIS and CLGS\_D obtain the
best results also with the DepthTrack test data thus verifying their
SotA performance in RGBD tracking;
{\bf 2)} as expected the long-term trackers, both RGB and RGBD, obtain
better F-scores than the short-term trackers;
{\bf 3)} the SotA performance numbers on DepthTrack are substantially lower
than with the CDTB dataset of VOT, for example, the VOT winner
ATCAIS obtains 0.702 with CDTB but only 0.476/0.483
with DepthTrack;
{\bf 4)} the new proposed baseline, DeT, which is trained with
generated depth data, wins on the both evaluation protocols and
obtains +12\% better F-score than the second best. At the same time the DeT
trackers has no long-term tracking capabilities and runs substantially faster than
the SotA RGBD trackers.

Our results verify that deep learning-based trackers, such as the proposed
DeT, provide
better accuracy than RGB trackers or ad hoc RGBD trackers on RGBD datasets,
but they need to be pre-trained with real RGBD data (D is generated in our experiments)
and fine-tuned using dataset specific training data.

\begin{figure*}[t]
\centering
\includegraphics[width=\linewidth]{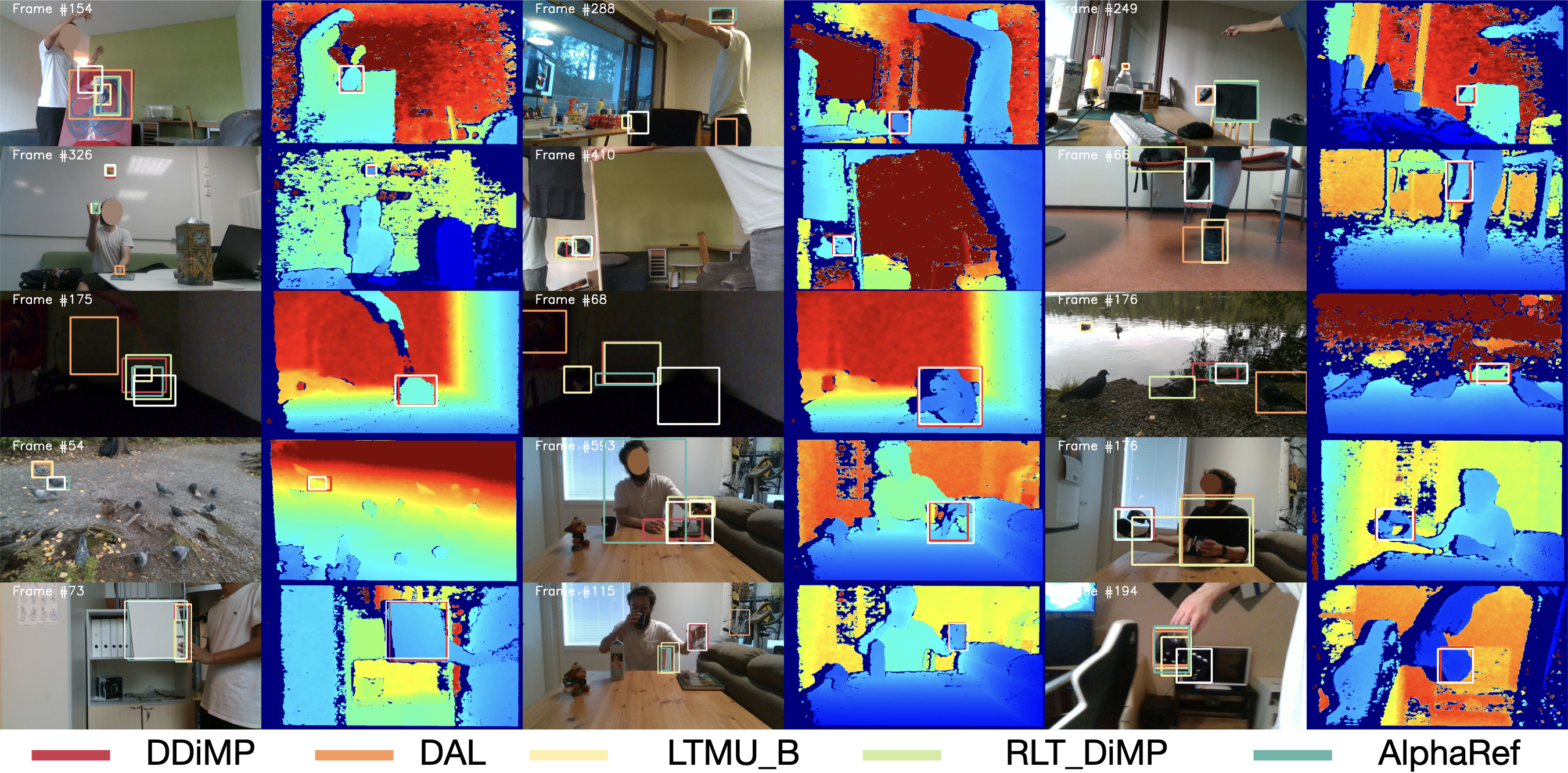}
\caption{
Difficult examples from the DepthTrack test set (targets marked with red boxes in depth images) :
similar background color (1st row);
multiple similar objects (2nd row);
dark scene (3rd row);
deformable (4th row); 
quickly rotating target (5th row) . In addition to the existing
baselines the DeT output is marked by white boxes and it is the only tracker
that succeeded on all example sequences.}
\label{fig:qualitative-evaluation}
\end{figure*}

\paragraph{Qualitative results.}
Fig.~\ref{fig:qualitative-evaluation} shows example tracks of
certain representative trackers and the proposed depth-trained
DeT tracker. With similar target and background color the
RGBD trackers perform clearly better than RGB trackers. In
addition, the depth cue helps to track the correct object in
the presence of multiple similar objects. On the other
hand, the depth-based trackers
have difficulties to track objects that rotate or move fast
in the depth direction. This is partly due to the reason that
the existing RGBD trackers extract features learned from RGB
data rather than from the both RGB and D channels. The problem can
be compensated by learning a ``depth backbone'' from depth
data as was done for the DeT which succeeded
on all the example sequences.

\begin{figure*}[h]
    \centering
    \includegraphics[width=0.85\linewidth]{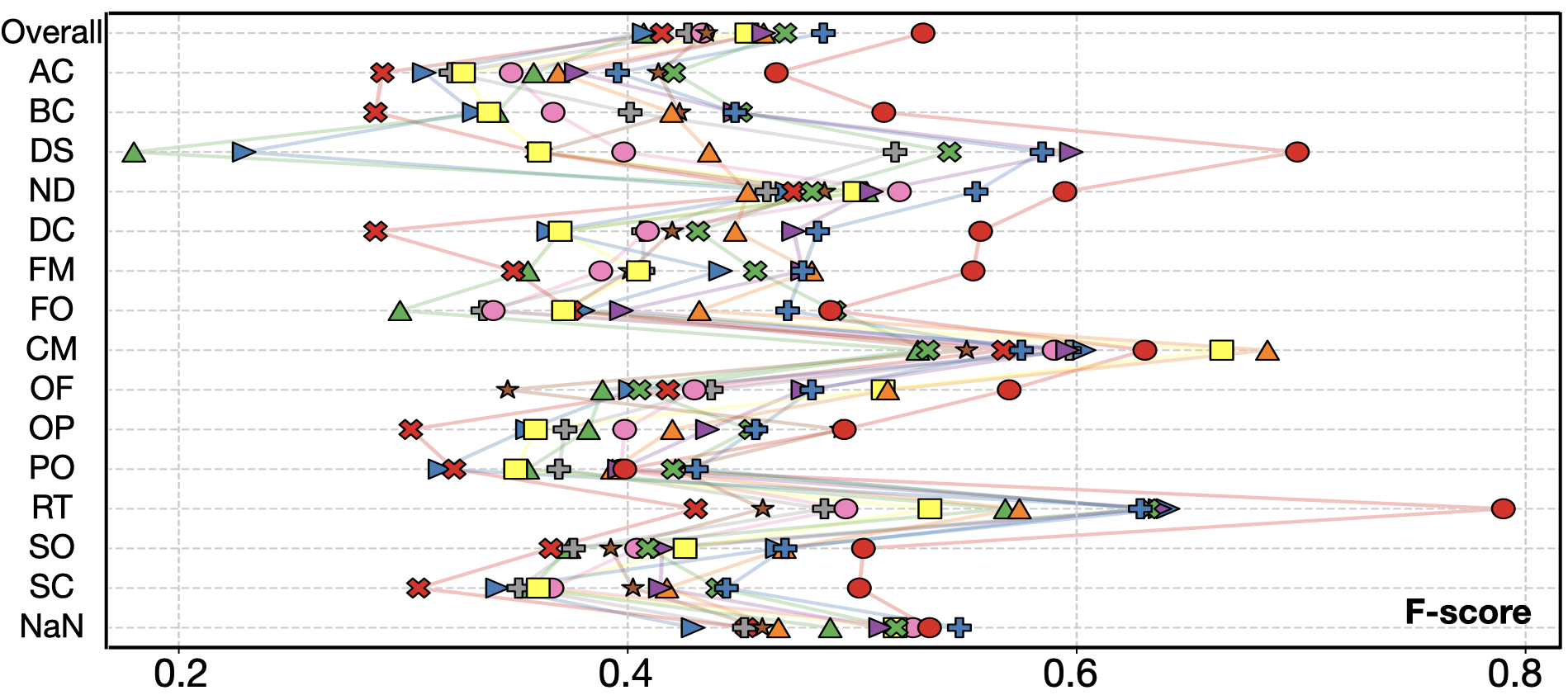}
    \caption{Optimal F-Score average over overlap thresholds for the visual attributes.
    \yan{The results of the best 10 evaluated trackers and the proposed DeT tracker and its master tracker DiMP50 are plotted
    (see Fig.~\ref{fig:fig_trackers_pr_re_fscore_final} for tracker markers).} 
    The abbreviation for each attribute is as follows: aspect change (AC), background clutter (BC), depth change (DC), fast motion (FM), dark scene (DS), full occlusion (FO), camera motion (CM), non-rigid deformation (ND), out of plane (OP), out of frame (OF), partial occlusion (PO), reflective targets (RT), size change (SC), similar objects (SO) and unassigned (NaN).}
    \label{fig:attribute_fscores}
\end{figure*}

\paragraph{Attribute-based performance analysis.}
\yan{Attribute specific F-scores for the best 10 tested trackers and the proposed DeT and the DiMP50 are shown in Fig.~\ref{fig:attribute_fscores}.}
The F-scores of the best VOT-RGBD 2020 tracker, ATCAIS, are clearly lower for the DepthTrack sequences. 
Interestingly, 
\yan{DeT wins in 11 of them, 
DDiMP (VOT-RGBD 2020 challenge second) in 2 and 
the remaining two are RLT\_DiMP and ATCAIS (VOT RGBD 2020 winner).}
It is worth noting that the long-term DDiMP 
handles target loss occasions, \eg \textit{partial occlusion} and \textit{out of frame}. While our DeT outperforms other trackers on most depth-related attributes, \eg \textit{dark scene} and \textit{depth change}.
Clearly the superior performance of RGBD over RGB is evident and using D training data with DeT makes it the most successful in long-term tracking evaluation even though it is a short-term tracker.

\paragraph{Computation times.}
The tracking speeds are reported in Table~\ref{tbl:experiments_on_DeTrack}.
The most important result of the speed comparison is that the
second best RGBD trackers, DDiMP and ATCAIS, only achieve the speed of 4.77 \textit{fps} and 1.32 \textit{fps} due to ad hoc depth processing.
On the other hand, the winning DeT tracker that is a
straightforward deep RGBD tracker architecture runs real-time.

\begin{table}[h]
\caption{Comparison of the original RGB trackers and their DeT variants.}
\label{tab:ablation_base}
\resizebox{1.0\linewidth}{!}{
\begin{tabular}{l r r r | r r r}
    \toprule
     Tracker & \multicolumn{3}{c|}{Sequence-based} & \multicolumn{3}{c}{Frame-based} \\
        ~    & Pr & Re & F-score & Pr & Re & F-score \\ 
        \midrule
        DiMP50~\cite{bhat2019learning} & 0.396 & 0.415 & 0.405 & 0.387 & 0.403 & 0.395 \\ 
        DeT-DiMP50-Max & 0.560 & 0.506 & 0.532 & 0.523 & 0.479 & 0.500 \\ 
        \midrule
        ATOM~\cite{2019ATOM} & 0.354 & 0.371 & 0.363 & 0.329 & 0.343 & 0.336 \\
        DeT-ATOM-Max & 0.457 & 0.423 & 0.440 & 0.438 & 0.414 & 0.426 \\
        \bottomrule
\end{tabular}
}
\end{table}

\begin{table}[h]
\caption{Comparison of the ``master'' tracker DiMP50 using only the depth channel input (depth-only tracking).
\yan{DiMP50(RGB) denotes the original RGB DiMP50 using the RGB input.
DiMP50(D) denotes the RGB trained DiMP50 using only depth input.
DeT-DiMP50(D) denotes the depth trained DeT with only depth input.} 
}
\label{tab:depth_only}
\resizebox{1.0\linewidth}{!}{
    \begin{tabular}{l r r r | r r r}
        \toprule
         Tracker & \multicolumn{3}{c|}{Sequence-based} & \multicolumn{3}{c}{Frame-based}  \\
            ~    & Pr & Re & F-score & Pr & Re & F-score\\ 
            \midrule
         DiMP50 (RGB)   & 0.396 & 0.415 & 0.405 & 0.387 & 0.403 & 0.395\\
         DiMP50 (D)     & 0.249 & 0.176 & 0.206 & 0.218 & 0.193 & 0.205\\
         DeT-DiMP50 (D)   & 0.412 & 0.377 & 0.394 & 0.423 & 0.400 & 0.411 \\
         \bottomrule
    \end{tabular}
}
\end{table}

\subsection{DeT Ablation Study}
\paragraph{Master trackers.}
As discussed in Sect.~\ref{sec:our_tracker} 
the \yan{DeT} can be attached to master trackers that expect such feature extraction pipelines. 
The most straightforward ``masters'' for DeT are  ATOM~\cite{2019ATOM} and DiMP50~\cite{bhat2019learning} which use the ResNet-18 and ResNet-50 RGB backbones in their original implementation. 
We compared the original RGB trackers to their DeT variants using the
depth-trained depth pathway for the D channel. 
The results are in Table~\ref{tab:ablation_base} that shows a clear
7.7\% improvement to original {ATOM} and 10.5\% to {DiMP50} in terms of F-score.

\paragraph{Depth cue.}
To further verify that superior performance of DeT is due to better
depth features learned by training with depth data, we compared 
\yan{the RGB trained depth-only DiMP {\em DiMP50(D)} and the depth trained DeT taking only depth input {\em DeT-DiMP50(D)}.}
The results are in Table~\ref{tab:depth_only}. 
\yan{DeT-DiMP50(D) is clearly superior to the DiMP50(D) and rather surprisingly the depth-only DeT-DiMP50(D) is almost on par with the highly optimized original RGB DiMP50.}

\paragraph{RGB and D feature pooling.}
In Sect.~\ref{sec:our_tracker} we selected the max pooling as the
RGB and D feature merging operation. In Table~\ref{tab:pooling} the results
are shown for the three different pooling layers: 1) the convolutional layer (``-MC''), 2) element-wise maximum (``-Max'') and 3) mean (``-Mean'').
All the pooling layers performed well as compared to using RGB features
for the D channel. There is only a small difference between
mean and max pooling, but we selected max pooling since DiMP performed
better than ATOM and was used in the method comparison experiment.

\begin{table}[h]
\caption{Comparison of the different pooling operations in the DeT feature merging layer.}
\label{tab:pooling}
\resizebox{1.0\linewidth}{!}{
\begin{tabular}{l|c c c | c c c}
    \toprule
     Tracker & \multicolumn{3}{c|}{Sequence-based} & \multicolumn{3}{c}{Frame-based}  \\
        ~    & Pr & Re & F-score & Pr & Re & F-score\\ 
        \midrule
      DiMP50          & 0.396 & 0.415 & 0.405 & 0.387 & 0.403 & 0.395\\
      DeT-DiMP50-MC   & 0.512 & 0.482 & 0.496 & 0.495 & 0.469 & 0.482 \\ 
      DeT-DiMP50-Max  & \bf{0.560} & 0.506 & \bf{0.532} & \bf{0.523} & 0.479 & \bf{0.500} \\
      DeT-DiMP50-Mean & 0.540 & \bf{0.520} & 0.530 & 0.508 & \bf{0.486} & 0.497 \\
      \midrule
      ATOM          & 0.354 & 0.371 & 0.363 & 0.329 & 0.343 & 0.336 \\
      DeT-ATOM-MC   & 0.472 & 0.423 & 0.446 & 0.420 & 0.400 & 0.411 \\ 
      DeT-ATOM-Max  & 0.457 & 0.423 & 0.440 & \bf{0.438} & 0.414 & 0.426 \\ 
      DeT-ATOM-Mean & \bf{0.479} & \bf{0.436} & \bf{0.457} & 0.435 & \bf{0.421} & \bf{0.428} \\
      \bottomrule
\end{tabular}
}
\end{table}
\begin{table}[h]
\caption{Cross-dataset experiment with the CDTB dataset. Trackers were trained using the DepthTrack data and no CDTB sequences were used to fine-tune the trackers or optimize their parameters.}
\label{tab:ablation_CDTB}
\resizebox{1.0\linewidth}{!}{
\begin{tabular}{l|c c c | c c c}
    \toprule
     Tracker & \multicolumn{3}{c|}{Sequence-based} & \multicolumn{3}{c}{Frame-based}  \\
        ~    & Pr & Re & F-score & Pr & Re & F-score\\ 
        \midrule
      DiMP50          & 0.549 & 0.555 & 0.552 & 0.546 & 0.549 & 0.547 \\
      DeT-DiMP50-MC   & 0.631 & 0.621 & 0.626 & 0.622 & 0.624 & 0.623 \\
      DeT-DiMP50-Max  & 0.651 & 0.633 & 0.642 & 0.626 & 0.621 & 0.624 \\ 
      DeT-DiMP50-Mean & \bf{0.674} & \bf{0.642} & \bf{0.657} & \bf{0.664} & \bf{0.644} & \bf{0.654} \\
      \midrule
      ATOM          & 0.548 & 0.536 & 0.542 & 0.541 & 0.537 & 0.539 \\ 
      DeT-ATOM-MC   & 0.582 & 0.567 & 0.574 & 0.572 & 0.574 & 0.573 \\
      DeT-ATOM-Max  & 0.583 & \bf{0.587} & \bf{0.585} & 0.575 & 0.583 & 0.579 \\
      DeT-ATOM-Mean & \bf{0.583} & 0.574 & 0.578 & \bf{0.589} & \bf{0.585} & \bf{0.587} \\
      \bottomrule
\end{tabular}
}
\end{table}
%



%

\paragraph{Cross-dataset evaluation.}
In order to verify that our findings from the DepthTrack experiments
are valid we compared DiMP and DeT architectures on
the CDTB dataset~\cite{Lukezic_2019_ICCV} of VOT-RGBD 2019 and 2020 and
without using any CDTB data in training (cross-dataset).
The results are shown
in Table~\ref{tab:ablation_CDTB} and they verify all findings except that
mean pooling performed better with the CDTB data.


\section{Conclusion}

In this work, we introduced a new dataset for RGBD tracking.
To the authors' best knowledge, the proposed DepthTrack is the largest
and most diverse RGBD benchmark and the first to provide a
separate training set for deep RGBD trackers. Our work is
justified by the lack of public datasets for RGBD tracking despite that
RGBD sensors have been available in consumer electronics for many years and
 are popular in robotics. RGB tracking has dominated
research in the field and the power of the depth cue for tracking
has remained unknown. In this work, we trained the first RGBD tracker
fully with RGBD data so that the depth pathway was trained 
with depth maps from scratch. The proposed DeT tracker was pre-trained with
generated RGBD sequences and then fine-tuned with the DepthTrack
training set. In all
experiments the DeT obtained the best F-score indicating that
depth plus RGB data trained trackers can finally unveil the
power of the depth cue for RGBD tracking.

{\small
\bibliographystyle{ieee_fullname}
\bibliography{egbib}
}

\end{document}